\documentclass[conference]{IEEEtran}
\IEEEoverridecommandlockouts
\usepackage{cite}
\usepackage{amsmath,amssymb,amsfonts}
\usepackage{algorithmic}
\usepackage{graphicx}
\usepackage{textcomp}
\usepackage{xcolor}
\usepackage{multirow}
\usepackage{booktabs}
\usepackage{bbding}
\usepackage{hyperref}
\hypersetup{colorlinks=true,
	linkcolor=black,
	anchorcolor=black,
	citecolor=black}
\def\BibTeX{{\rm B\kern-.05em{\sc i\kern-.025em b}\kern-.08em
    T\kern-.1667em\lower.7ex\hbox{E}\kern-.125emX}}
\begin{document}

\title{VADMamba: Exploring State Space Models for Fast Video Anomaly Detection
\thanks{\IEEEauthorrefmark{1}Corresponding author.} 
\thanks{This work was supported by Natural Science Foundation of Shaanxi Province, China(2024JC-ZDXM-35, 2024JC-YBMS-458, 2024JC-YBMS-573), National Natural Science Foundation of China (No.52275511) and Young Talent Fund of Association for Science and Technology in Shaanxi, China (20240146).}}

\author{Jiahao Lyu, Minghua Zhao\IEEEauthorrefmark{1}, Jing Hu, Xuewen Huang, Yifei Chen, Shuangli Du \\
School of Computer Science and Engineering, Xi'an University of Technology, Xi'an, China \\
zhaominghua@xaut.edu.cn}


\maketitle

\begin{abstract}

Video anomaly detection (VAD) methods are mostly CNN-based or Transformer-based, achieving impressive results, but the focus on detection accuracy often comes at the expense of inference speed. The emergence of state space models in computer vision, exemplified by the Mamba model, demonstrates improved computational efficiency through selective scans and showcases the great potential for long-range modeling. Our study pioneers the application of Mamba to VAD, dubbed VADMamba, which is based on multi-task learning for frame prediction and optical flow reconstruction. Specifically, we propose the VQ-Mamba Unet (VQ-MaU) framework, which incorporates a Vector Quantization (VQ) layer and Mamba-based Non-negative Visual State Space (NVSS) block. Furthermore, two individual VQ-MaU networks separately predict frames and reconstruct corresponding optical flows, further boosting accuracy through a clip-level fusion evaluation strategy. Experimental results validate the efficacy of the proposed VADMamba across three benchmark datasets, demonstrating superior performance in inference speed compared to previous work. Code is available at https://github.com/jLooo/VADMamba.

\end{abstract}


\section{Introduction}
\label{sec:intro}

Detecting anomalous events from surveillance videos manually is extremely tedious and time-consuming \cite{liu2024generalized}, and achieving fast discrimination of anomalies is even more challenging for the uninitiated. The demand for the detection of massive videos has greatly contributed to the intelligent development of video anomaly detection (VAD).
\begin{figure}
	\centering
	\includegraphics[width=\linewidth]{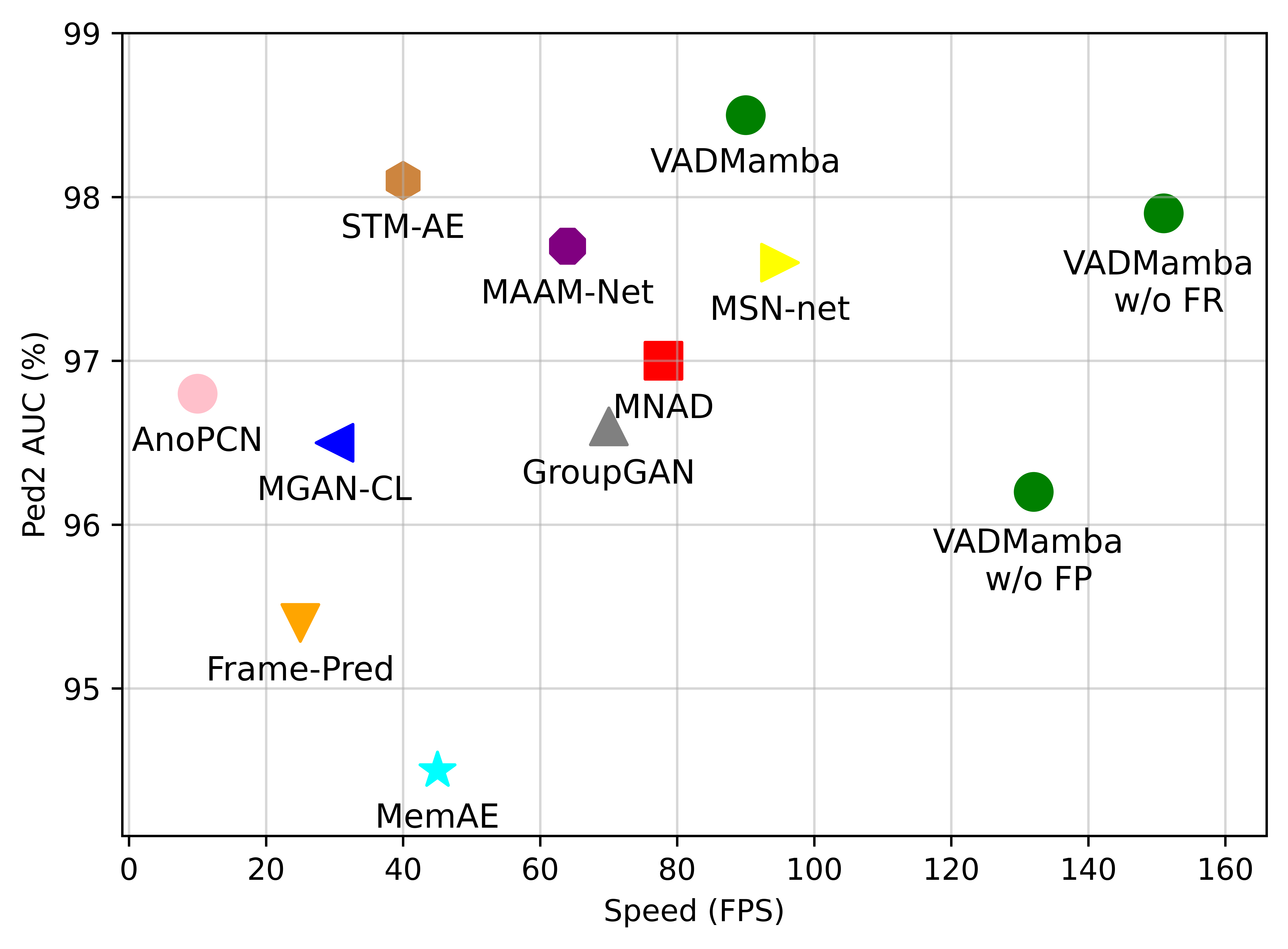}
	\caption{Comparison of inference speed (FPS) and frame-level AUC (\%) on Ped2. VADMamba demonstrates state-of-the-art performance in terms of FPS.}
	\label{fig:1}
\end{figure}

Traditional studies were based on handicraft features such as Local Binary Patterns \cite{hu2018squirrel} and Histogram of Gradients \cite{hasan2016learning} to replace manual, but such methods are limited by a priori knowledge and have poor performance. With the continuous development of Deep Learning (DL) and Computer Vision, the VAD studies based on Convolutional Neural Networks (CNNs) and Transformer have shown great success. However, there are native challenges due to the sparse and discrete distribution of the anomaly samples and the high cost of manual annotation \cite{chandrakala2023anomaly}, most of the VADs are based on One-Class Classification (OCC) \cite{yang2023video,lyu2024appearance} to detect anomalies. OCC is defined as learning normal samples only in the training phase, generating normal latent representations, whereas in the inference phase, any external feature representation is then defined as an anomaly \cite{liu2024generalized}. Therefore, the focus of previous studies is on how to improve the discrimination between normal and abnormal samples. In DL-based VAD, reconstruction-based \cite{gong2019memorizing,park2020learning,singh2024attention,qiu2024video} focuses on modeling spatial features, while prediction-based \cite{liu2018future,liu2023msn,cheng2023spatial,lyu2024bidirectional} focuses on modeling temporal features, and another very few methods \cite{yu2020cloze, hu2022detecting} use visual cloze to capture high-level semantics to perform VAD, such that different methods have their advantages and disadvantages for different types of anomalies. Consequently, some works have combined reconstruction and prediction to develop the hybrid VAD \cite{ye2019anopcn,liu2022learning,li2023multi,wang2023memory,sun2024dual,liang2024c,ijcai2024p96} to obtain high performance.

\begin{figure*}[ht]
	\centering
	\includegraphics[width=\linewidth]{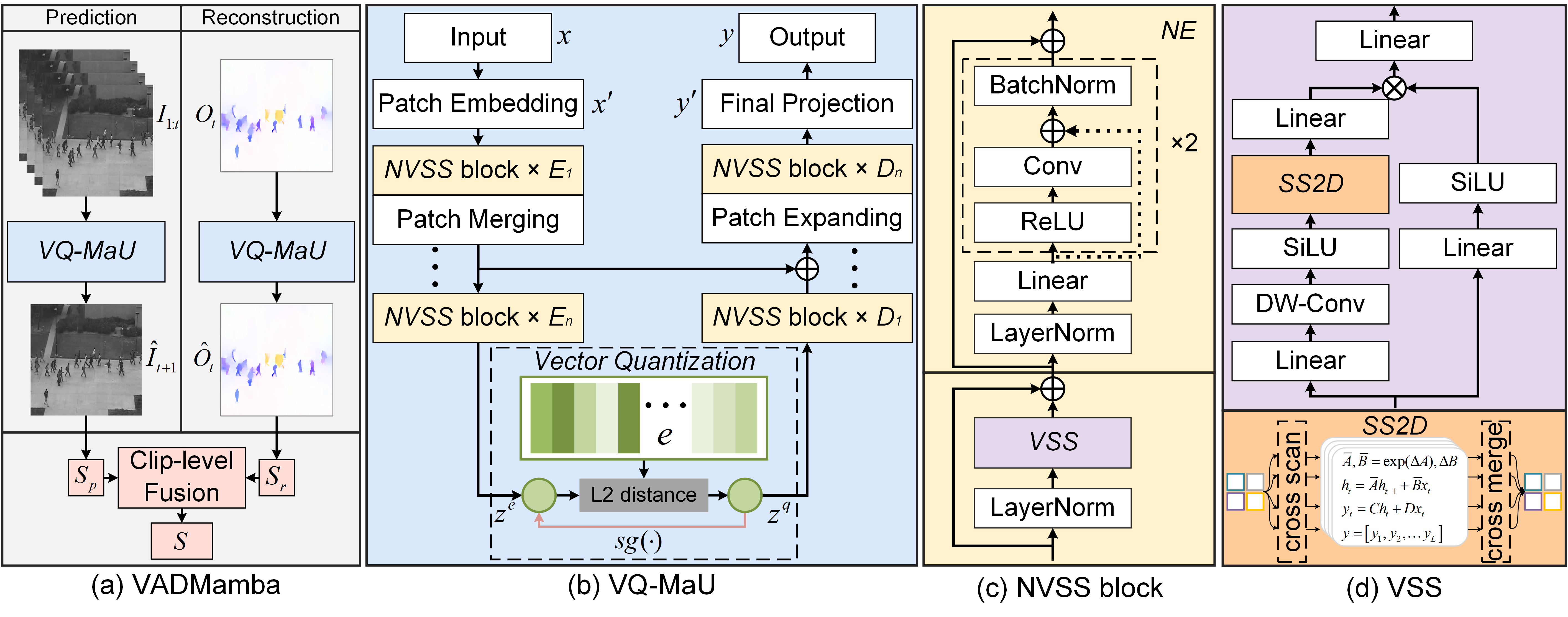}
	\caption{Overview of the proposed VADMamba. (a) The training and inference process of VADMamba. (b) The framework of the proposed VQ-MaU. (c) Non-negative Vision State Space block. The dashed line indicates that addition is used in the second loop. (d) Vision State-Space (VSS) with SS2D.}
	\label{fig:2.1}
\end{figure*}

Deep CNNs can extract multi-scale local spatial features through various convolving kernels, and can also obtain context features, but are limited by local receptive fields and lack sufficiently expressive in capturing global spatial features or longer temporal features. Vision Transformers (ViTs) \cite{dosovitskiy2020image} converts the image into a sequence of image patches, and ViTs are easier to extract long-range dependencies than CNNs. However, as the sequence length grows, the model requires more memory costs and computational resources. To solve the above problems, state space models (SSMs) \cite{guefficiently} have been widely studied. In particular, the latest work Mamba \cite{gu2023mamba,daotransformers} has achieved Transformers-like long-range modeling capability and linear scalability of sequence length by optimizing the structure SSM and proposing the selective structure SSM combined with hardware-aware algorithms. Many recent studies in the field of computer vision have pioneered introducing Mamba \cite{liu2024vmamba,zhuvision,hemambaad,ruan2024vm,lu2024videomambapro}. To our knowledge, our work is the first Mamba-based VAD that achieves high accuracy and fast inference speed. More specifically, we introduce VQ-Mamba Unet (VQ-MaU), which designs vector quantization at the bottleneck to achieve discrimination enhancement among different samples by compression of normal features. Then, we propose a Non-negative Vision State Space (NVSS) basis block based on SSMs to accelerate the speed of different feature aggregation and model convergence through pre-activation. Finally, we implement a hybrid VAD that aggregates frame prediction and optical flow reconstruction. In the inference phase, VADMamba achieves higher performance through clip-level fusion evaluation strategy. Our main contributions are:

\begin{itemize}
	\item We propose VADMamba, the first work to explore and apply Mamba to solve VAD, by introducing frame prediction and optical flow reconstruction to achieve high-precision hybrid detection.
	\item We introduce VQ-MaU, which enables the retrieval and discrimination of feature representations from compressed normal features via the VQ layer, and construct NVSS blocks to enhance feature aggregation speed. 
	\item We propose that the clip-level fusion evaluation strategy effectively exploits the advantages of different input features, thereby improving detection performance.
	\item The effectiveness of the proposed VADMamba is validated on three benchmark datasets, demonstrating excellent accuracy while achieving a fast inference speed.
\end{itemize}

\section{PROPOSED METHOD}
\label{sec:METHOD}

The training and inference stage of VADMamba, as illustrated in Fig.~\ref{fig:2.1}(a), involves frame prediction, optical flow reconstruction corresponding to the future frame, and anomaly score calculation. \textbf{(1) Frame Prediction (FP):} The model takes $t$ consecutive input frames, $I_{1:t}$, to predict the next frame, $I_{t+1}$. \textbf{(2) Flow Reconstruction (FR):} Based on the predicted frame, $I_{t+1}$, the optical flow $O_{t}$ is reconstructed. \textbf{(3) Inference stage:} The anomaly scores $S_{p}$ and $S_{r}$, obtained from FP and FR, are combined through a clip-level fusion evaluation strategy to produce a hybrid score $S$.

For the two tasks above, FP is trained first, followed by FR, which is trained using the best-performing FP model. FR focuses on reconstructing the optical flow corresponding to the predicted frames from FP to achieve optimal hybrid detection performance. Parallel training was avoided because the two tasks converge at different speeds, making simultaneous optimization difficult and leading to suboptimal hybrid detection performance. In the following subsections, we describe in detail all the components of our proposed VADMamba.

\noindent{\textbf{VQ-MaU.}} The proposed VQ-MaU framework is shown in Fig.~\ref{fig:2.1}(b), which includes the Patch Embedding layer, NVSS block-based encoder and decoder, vector quantization layer, and Final Projection layer. VQ-MaU designs a symmetric Unet structure with the skip connection. In our experiments, the depth $n$ of the codec is 4. In the encoder stage, first, the Patch Embedding layer will segment the input image $x \in \mathbb{R} ^{H\times W \times C}$ into a non-overlapping patch of size $4\times 4$, and map the image to 64 channels to generate the embedded image ${x}' \in \mathbb{R} ^{H/4\times W/4 \times 64}$. Then, $E_{n}$ NVSS blocks are used for each encoder stage, input ${x}'$ to first NVSS, and perform Patch Merging down-sampling \cite{liu2021swin} (except for the last encoder stage), and finally the encoder feature $z_{e}$ is input to VQ. In the decoder stage, first, the representation features $z_{q}$ by the VQ compression are input to the decoder NVSS block, and then, the features are coupled with skip feature stage by stage, again using $D_{n}$ NVSS blocks per decoder stage, and Patch Expanding up-sampling \cite{cao2022swin} before each NVSS block (except for the first decoder stage). Finally, the Final Projection layer restores the feature ${y}'$ to $y$ of the same size as the input $x$ to match prediction or reconstruction. For the skip connection, we directly use the element-by-element addition operation.

\noindent{\textbf{Non-negative Vision State Space.}} As shown in Fig.~\ref{fig:2.1}(c), the proposed NVSS block consists of two parts: (1) the Vanilla VSS \cite{liu2024vmamba} block uses depth-wise convolution for high-level feature extraction, and the other part uses linear mapping and activation functions to compute gating signals. The core is the SS2D (2D-Selective-Scan) block in Fig.~\ref{fig:2.1}(d), which establishes a global receptive field through complementary traversal paths, enabling each pixel to efficiently gather information from all others across multiple directions. (2) Non-negative Enhanced (NE) module is proposed to optimize the model for mode-specific feature aggregation and accelerate convergence. First, the features are normalized by Layernorm and linear mapping operations, and then introduce an unconventional pre-activation  \cite{he2016identity}, i.e., the structural design transforms to ReLU$\rightarrow$Conv$\rightarrow$BN, which inputs the non-negative features into the convolution to alleviate the gradient vanishing and reduce the overfitting.

\noindent{\textbf{Vector Quantization.}} To reduce feature dimensions and provide high-quality hidden features, as well as common normal data with small errors and rare abnormal data with large errors, we inserted vector quantization (VQ) \cite{van2017neural} in the bottleneck to compress the features, which maximises the retention of essential information without overfitting. The VQ module defines the codebook vector $e \in \mathbb{R}^{K \times d}$, where $K$ is the size of the codebook and $d$ is the dimension of codes. The vector-quantized representation $z^{q}$ is determined by looking up the codebook vector closest to the input features $z^{e}(x) \in \mathbb{R}^{h \times w \times d}$ in terms of the L2 distance:
\begin{equation}
z_{k}^{q}=argmin_{j}\left \| z^{e}(x)-e_{j} \right \|_{2},
\end{equation}
where $j \in \{1,2,\dots,K\}$ and $e_{j}$ represents the $j^{th}$ item.

The training objective of VQ is defined as follows:
\begin{equation}
\mathcal{L}_{vq}=\left\|sg\left(z^{e}(x)\right)-z^{q}\right\|_{2}+\beta\left\|z^{e}(x)-s g\left(z^{q}\right)\right\|_{2},
\end{equation}
where $sg(\cdot)$ represents the stop-gradient operator defined as the identity in the forward computation and $\beta$ set to 0.25.

\noindent{\textbf{Training Loss.}} The overall loss function of FP consists of prediction loss $\mathcal{L}_{p}$, vq loss $\mathcal{L}_{vq}$, and gradient loss $\mathcal{L}_{gd}$: $\mathcal{L}_{FP}=\mathcal{L}_{p}+\mathcal{L}_{vq}+\mathcal{L}_{gd}$. 
\begin{equation}\label{eq1}
\mathcal{L}_{p}=\left\|I_{t+1}-\hat{I}_{t+1}\right\|_{2},
\end{equation} 
\begin{equation}\label{eq:11}
\begin{gathered}
\mathcal{L}_{gd}=\sum_{i,j} \left \| \left | I_{i,j}-I_{i-1,j} \right | - \left | \hat{I}_{i,j}-\hat{I}_{i-1,j} \right | \right \|_{1} + \\
\left \| \left | I_{i,j-1}-I_{i,j} \right | -\left | \hat{I}_{i,j-1}-\hat{I}_{i,j} \right | \right \|_{1},  
\end{gathered}
\end{equation}
where $i$ and $j$ represent the spatial index of the frame.

The overall loss function of FR consists of reconstruction loss $\mathcal{L}_{r}$, vq loss $\mathcal{L}_{vq}$, similarity loss $\mathcal{L}_{sim}$, and additional motion difference loss $\mathcal{L}_{md}$ for motion features: $\mathcal{L}_{FR}=L_{r}+\mathcal{L}_{vq}+\mathcal{L}_{sim}+0.01\mathcal{L}_{md}$.
\begin{equation}\label{eq12}
\mathcal{L}_{r}=\left\|O_{t}-\hat{O}_{t}\right\|_{2},
\end{equation} 
\begin{equation}\label{eq3}
\mathcal{L}_{sim}=1-\operatorname{SSIM}(O_{t}, \hat{O}_{t}),
\end{equation}
where SSIM denotes the Structural Similarity Index Measure to compute the similarity between the real flow $O_{t}$ and the reconstructed flow $\hat{O}_{t}$.
\begin{equation}\label{eq13}
\mathcal{L}_{md}=\sqrt{\|\| M_{t}\left\|_{2}-\right\| \hat{M}_{t}\left\|_{2}\right\|^{2}+\varepsilon^{2}},
\end{equation} 
where $M_{t}$ is the motion difference between $O_{t}$ and $O_{t-1}$, similarly, $\hat{M}_{t}$ is the difference between $\hat{O}_{t}$ and $O_{t-1}$, $\varepsilon$ is a small positive constant set to 0.001.

\noindent{\textbf{Inference Score.}} Following \cite{liu2018future}, we use the Peak Signal to Noise Ratio (PSNR) generated by the frame prediction error and the optical flow reconstruction error as the anomaly scores: $S_{p}=\text{PSNR}(I_{t+1}, \hat{I}_{t+1})$ and $S_{r}=\text{PSNR}(O_{t}, \hat{O}_{t})$. Then we normalize PSNR to $[0, 1]$ range by applying \eqref{eq:16}:
\begin{equation}\label{eq:16}
S(I_{t})=\frac{\text{PSNR}(I_{t},\hat{I}_{t})-\text{min}_{t}(\text{PSNR}(I_{t},\hat{I}_{t}))}{\text{max}_{t}(\text{PSNR}(I_{t},\hat{I}_{t}))-\text{min}_{t}(\text{PSNR}(I_{t},\hat{I}_{t}))} .
\end{equation}


Due to the inherent temporal continuity in videos, we apply a Gaussian filter to smooth the error scores. Unlike previous works that rely on frame-level fusion, we propose a clip-level fusion strategy that evaluates video segments rather than individual frames. This evaluation strategy selects the superior score between the frame scores $S_{p}$ and optical flow scores $S_{r}$ based on the AUC of each video clip and combines the selected clip scores to produce the final hybrid score $S$. Fusing the evaluation of frame and optical flow proved more effective than treating these tasks separately.
\begin{equation}\label{eq:17}
S^{(i)}=\left\{\begin{matrix} 
S_p^{(i)}, & \text{if } \text{AUC}(S_p^{(i)}) \geq \text{AUC}(S_r^{(i)}), \\  
S_r^{(i)}, & \text{otherwise.}  
\end{matrix}\right. 
\end{equation}
where $i \in \{1,2,\dots,N\}$, $N$ is the number of video clips and $S^{(i)}$ indicates all the scores of the $i$-th clips.

\section{EXPERIMENTS}
\label{sec:EXPERIMENTS}
\subsection{Implementation Details}\label{sec:31}
We conducted experiments on UCSD Ped2 \cite{sabokrou2015real}, CUHK Avenue \cite{lu2013abnormal} and ShanghaiTech (SHT) \cite{luo2017revisit}. Following the study \cite{park2020learning}, the Area Under ROC curve (AUC) is used as the evaluation metric. We resized each input frame to $[-1,1]$ intensity and resolution to $256\times 256$. The optical flow is obtained using Flownet 2.0 \cite{ilg2017flownet} with the same resolution. AdamW is used for model initialization with a learning rate of 2e-4. The input length $t$ of FP is 16. For different datasets, the number of NVSS blocks for $E_{n}$ and $D_{n}$ are as follows: In Ped2 and SHT, the $E_{n}$ and $D_{n}$ are \{1,1,1,1\}, and in Avenue are \{2,2,2,2\}. For FP and FR tasks, the NVSS block configuration is the same.

\subsection{Quantitative Comparison with Existing Methods}

We compared VADMamba with the state-of-the-art single-task methods \cite{liu2018future,gong2019memorizing,park2020learning,yu2020cloze,hu2022detecting,liu2023msn,cheng2023spatial,qiu2024video,yang2023video,singh2024attention} and hybrid methods \cite{ye2019anopcn,liu2022learning,li2023multi,wang2023memory,sun2024dual,liang2024c,ijcai2024p96}, and Table~\ref{tab:sota} shows the results of the frame-level AUC. In addition, we compared two single-task variants of VADMamba, 'VADMamba w/o FP' and 'VADMamba w/o FR', which were developed by removing FP and FR to formulate them. Note that the variant model results are all optimal under the same task, independent of the hybrid task. 
\begin{table}[ht]
	\centering
	\caption{Comparison of frame-level AUC (\%) with state-of-the-art methods. The top two results in each category are marked in \textbf{bold} and \underline{underline}.}
	\label{tab:sota}
	
	\begin{tabular}{c|l|c|c|c|c|c}
		\midrule
		& Method   &  Model   & Ped2 & Avenue & SHT & FPS\\ \midrule
		\multirow{12}{*}{\rotatebox{90}{Single}} 		
		& Frame-Pred \cite{liu2018future} & CNN    & 95.4 & 85.1   & 72.8 & 25 \\		
		& MemAE \cite{gong2019memorizing}   & CNN       & 94.1 & 83.3   & 71.2 & 45\\
		& MNAD  \cite{park2020learning}   & CNN      & 97.0 & 88.5   & 70.5 & 78\\
		& VEC \cite{yu2020cloze}			& CNN     & 97.3 & 90.2   & 74.8 & -\\	
		& TMAE	\cite{hu2022detecting} & ViT      & 94.1 & 89.8   & 71.4 & -\\
		& MSN-net \cite{liu2023msn}    & CNN    & 97.6 & 89.4   & 73.4 & 95\\
		& STGCN-FFP \cite{cheng2023spatial}  & CNN    & 96.9 & 88.4   & 73.7 & -\\
		& A2D-GAN \cite{singh2024attention} & GAN    & 97.4 & \textbf{91.0}   & \underline{74.2} & -\\
		& VADCL \cite{qiu2024video}  & ViT    & 92.2 & 86.2   & 73.8 & -\\
		& USTN-DSC  \cite{yang2023video}   & ViT   & \textbf{98.1} & \underline{89.9}   & 73.8 & -\\ 
		
		& VADMamba w/o FP & SSM  & 96.2 & 87.3     & 72.2  & 132\\ 
		& VADMamba w/o FR & SSM  & \underline{97.9} & 86.2     & \textbf{74.4} & 151\\ 
		\midrule 
		\multirow{8}{*}{\rotatebox{90}{Hybrid}}
		& AnoPCN \cite{ye2019anopcn}& CNN & 96.8 & 86.2   & 73.6 & 10\\
		& STM-AE \cite{liu2022learning}   & CNN      & \underline{98.1} & 89.8   & 73.8 & 40\\
		& MGAN-CL \cite{li2023multi}& GAN & 96.5 & 87.1   & 73.6 & 30\\
		& MAAM-Net \cite{wang2023memory} & CNN    & 97.7 & \underline{90.9}   & 71.3 & 64\\
		& GroupGAN \cite{sun2024dual}& GAN & 96.6 & 85.5   & 73.1 & 70\\
		& C$^{2}$Net  \cite{liang2024c} & GAN     & 98.0 & 87.5   & 71.4 & -\\ 
		& PDM-Net  \cite{ijcai2024p96} & CNN     & 97.7 & 88.1   & \underline{74.2} & -\\ 
		& VADMamba & SSM  & \textbf{98.5} & \textbf{91.5}     & \textbf{77.0} & 90\\ 
		\midrule
	\end{tabular}
\end{table}

It can be noticed that VADMamba achieves state-of-the-art performance on the three datasets due to the complementarity between the two tasks, and 'w/o FR' also achieves optimal performance on SHT. Interestingly, on the Avenue dataset, the hybrid task performs much better than the single task. It can be found that unlike Ped2 and SHT, optical flow reconstruction plays a decisive role in Avenue, which is due to factors such as the complex background of Avenue and the uneven illumination of the background between different clips, which leads to the inability of the VQ compressed features to form normal clusters, while the optical flow information avoids the background interference and improves the performance. It also further proves the advantage of clip-level fusion evaluation strategy. Using Ped2 as an example, the inference speed of the prediction, reconstruction, and hybrid model is 151, 132, and 90 FPS, respectively, including data processing and anomaly score calculation, which prove to outperform the same category.

\subsection{Visualization} 
\begin{figure}[h]
	\centering
	\includegraphics[width=\linewidth]{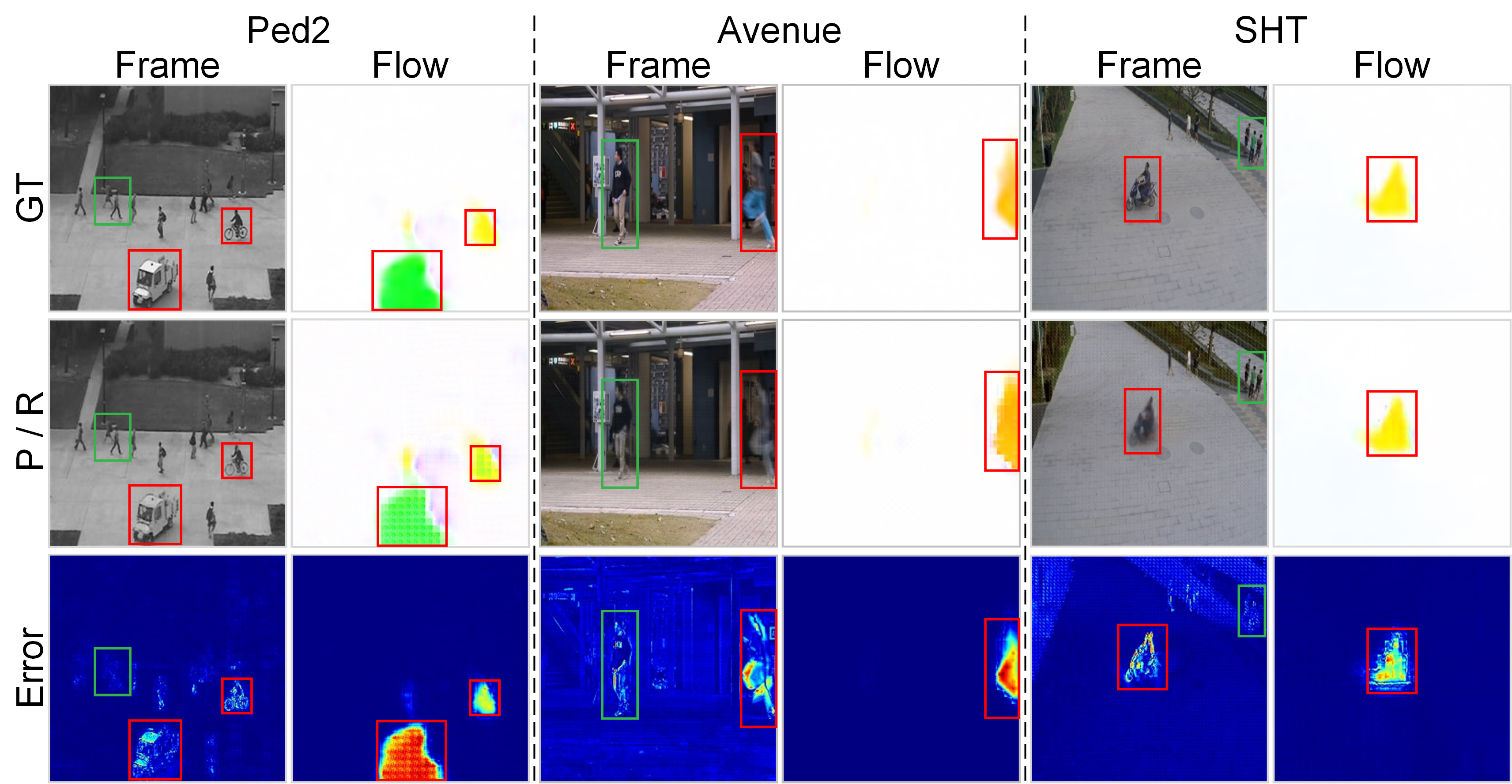}
	\caption{Visualization examples of FP and FR. From top to bottom, we show ground truth (GT), predicted frames (P), reconstructed optical flows (R), and error maps (Error). In the error map, brighter color indicates larger errors. The objects remarked with red/green borders are the anomaly/normal events.}
	\label{fig:3}
\end{figure}

Fig.~\ref{fig:3} presents two visualization results from three datasets, including the predicted frames and reconstructed optical flow for normal and abnormal events. In video frames, VADMamba predicted normal events well, while anomalies were successfully identified through deviations in the predictions. In the optical flow, the weak amplitude of the normal motion prevents the generation of normal optical flow, while the reconstruction of the abnormal optical flow takes significant errors. It can be observed that VADMamba performs exceptionally well in handling video anomalies, both in terms of video frames and optical flow. Specifically, the error of the reconstructed flow is independent of the background compared to the predicted frames, indicating that the optical flow mitigates the interference caused by the background in Avenue and SHT, thereby enhancing the overall detection performance of the method.

\begin{figure}[h]
	\centering
	\includegraphics[width=\linewidth]{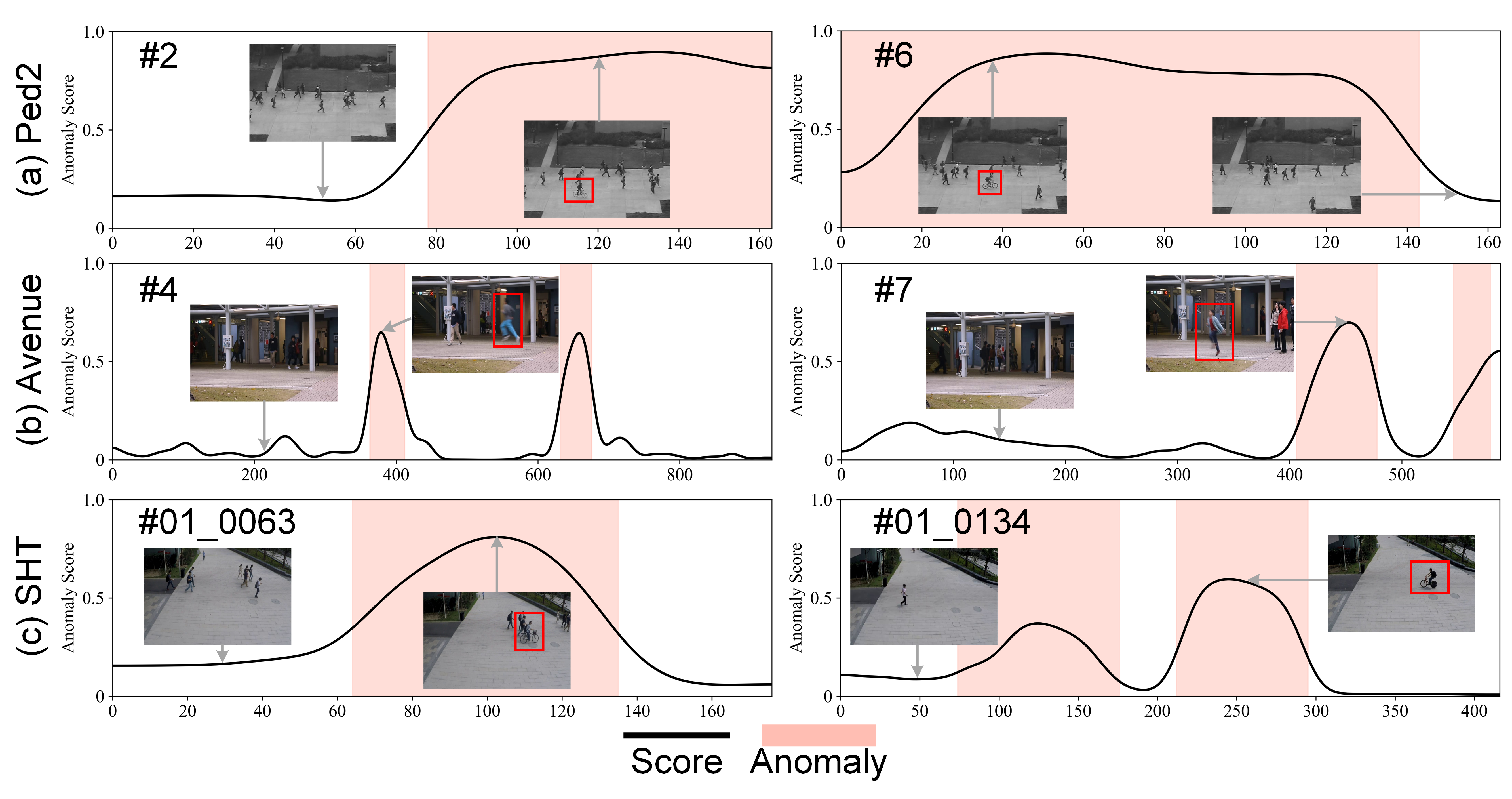}
	\caption{Anomaly score curves for six examples. Red regions indicate anomalous events, with larger values indicating a greater likelihood of anomalies.}
	\label{fig:4}
\end{figure}

Fig.~\ref{fig:4} further visualizes the anomaly curve from the two test video clips in the three datasets. The anomaly score is computed using an adaptive threshold. Each test video frame is classified by normalizing the anomaly score with the threshold, where a score near 0 indicates normal, and a score near 1 indicates abnormal. It is evident that VADMamba can distinguish between abnormal and normal very well. In the Ped2 \#2 and \#6 test videos, two continuous anomalous events (cycling) are detected. In the Avenue \#4 and \#7 test videos, for more discrete anomalous events (running, playing), VADMamba is also able to detect them. We can observe that the anomaly score rises extremely in the time dimension when an anomalous event occurs and there is no leakage of anomalous events. In the SHT \#01\_0063 and \#01\_0134 test videos, the abnormal cycling events in different directions are detected. This reflects VADMamba's strong capabilities in VAD, both in the temporal and spatial dimensions.

\subsection{Ablation Studies} 
\noindent{\textbf{Model components.}} To evaluate the effect of different components of VADMamba, the ablation study is performed on the Ped2 and Avenue datasets under three tasks. Table~\ref{tab:a1} reports the AUC results of the ablation experiments across different tasks. By separately adding the NE and VQ models, it was observed that VQ outperformed NE in most of the three tasks.  This is because, in prediction tasks, the diversity of consecutive frames allows VQ to capture various normal features, preventing overfitting.  On the other hand, the high similarity of features within one optical flow information makes it challenging to efficiently form normal clusters. For the NE module, optical flow information is easier to extract features from compared to complex frames. Ultimately, by incorporating both models, the proposed method achieved optimal performance across all three tasks, demonstrating the excellent performance of VADMamba.
\begin{table}[ht]
	\centering
	\caption{Component ablation study under three tasks.}
	\label{tab:a1}
	
	\begin{tabular}{cc|cc|cc|cc}
		\midrule
		\multirow{2}{*}{VQ} & \multirow{2}{*}{NE} & \multicolumn{2}{c|}{FP} & \multicolumn{2}{c|}{FR} & \multicolumn{2}{c}{MIX} \\
		&  & Ped2 & Avenue & Ped2 & Avenue & Ped2 & Avenue \\
		\midrule
		\XSolidBrush & \Checkmark   & 97.5 & 85.5 & 95.6 & 86.6  & 97.5  & 89.5   \\
		\Checkmark   & \XSolidBrush & 97.6 & 85.7 & 95.4 & 86.7  & 97.7  & 90.7   \\ 
		\Checkmark   & \Checkmark   & 97.9 & 86.2 & 96.2 & 87.3  & 98.5  & 91.5  \\
		\midrule
	\end{tabular}
	
\end{table}

\noindent{\textbf{Model variants.}} 
We compare three model variants for two tasks in terms of accuracy and parameter count, as shown in Table~\ref{tab:xx}. The evaluation assesses the performance of the FP and FR tasks under varying NVSS block configurations in VADMamba. Refer to Section~\ref{sec:31}, the NVSS blocks of $E_{n}$ and $D_{n}$ are identical, so a single configuration is used to represent both. Notably: (1) Increasing the number of NVSS blocks significantly raises the parameter count (up to 3$\times$), but does not always improve performance. On the Ped2 and SHT datasets, deeper networks tend to overfit, reducing detection accuracy. (2) Models with a moderate number of NVSS blocks achieve the best overall results, striking a balance between accuracy and computational efficiency. For instance, Model 2 (28M parameters), excels on Avenue (FP: 86.2\%, FR: 87.3\%) and maintains competitive performance across other datasets. (3) The lightweight Model 1 (14M parameters) performs well on simpler background-constructed datasets, such as Ped2 (FP: 97.9\%), but struggles with more complex datasets, highlighting its limitations in feature extraction. In summary, this study highlights a trade-off between model complexity and performance. While increasing the number of NVSS blocks enhances the model's representational power, excessive depth can lead to overfitting and inflate computational costs.

\begin{table}[h]
	\centering
	\caption{The different number of NVSS blocks for different datasets under two tasks.}
	\label{tab:xx}
	\begin{tabular}{ccccccc}
		\midrule
		Model & Blocks & Task & Ped2 & Avenue & SHT  & Params \\
		\midrule
		\multirow{2}{*}{1} & \multirow{2}{*}{\{1, 1, 1, 1\}} & FP & \textbf{97.9} & 85.7 & \textbf{74.4} & 14.84M    \\
		&         & FR & \textbf{96.2} & 86.2 &  \textbf{72.2} &  14.77M  \\
		\midrule
		\multirow{2}{*}{2} & \multirow{2}{*}{\{2, 2, 2, 2\}} & FP & 97.5  & \textbf{86.2} & 73.0 & 28.21M  \\
		&        & FR & 96.2  & \textbf{87.3} & 72.0    & 28.14M  \\
		
		\midrule
		\multirow{2}{*}{3} & \multirow{2}{*}{\{3, 3, 3, 3\}} & FP & 97.4 & 85.3 & 73.0  & 41.57M  \\
		&         & FR & 96.0 & 86.0   & 71.5 &  41.51M  \\
		
		\midrule
	\end{tabular}
\end{table}

\noindent{\textbf{Varying the number of input frames.}} We set different input frame numbers $t$ to measure the performance of the CNN-based MNAD \cite{park2020learning} and the proposed SSM-based VADMamba. Specifically, MNAD is effective only when dealing with shorter time series data, and as $t$ increases, the performance of the MNAD model gradually decreases, from 96.3\% to 95.3\%. This decreasing trend suggests that CNNs may suffer from performance degradation due to their inability to capture long-range dependencies when dealing with longer sequences effectively. In contrast, FP and MIX can better capture complex long-range dependencies. In particular, the performance of SSM of arbitrary length is better than or equal to MNAD under most conditions, which proves that this performance advantage is more significant as the time series lengthens. It is shown that SSM is not only able to effectively handle long-range dependency tasks but also able to maintain excellent performance in shorter-time dependency tasks. In particular, the MIX model can maintain excellent performance across different input frame lengths by integrating the advantages of prediction and reconstruction methods.

\begin{table}[h]
	\centering
	\caption{The AUC (\%) obtained by different methods for different $t$ on the Ped2 dataset. $\ast$ denotes AUC mentioned in the source paper.}
	\begin{tabular}{ccccc}
		\midrule
		Method & $t = 4$ & $t = 8$ & $t = 12$ & $t = 16$\\
		\midrule
		MNAD-P & 96.3 (97.0${\ast}$) & 96.0 & 95.9 & 95.3 \\
		FP & 96.3 & 96.7 & 97.3 & 97.9 \\
		MIX & 97.1 & 97.3 & 97.8 & 98.5 \\
		\midrule
	\end{tabular}
\end{table}

\section{Conclusion}
\label{sec:Conclusion}
We introduce VADMamba, the first mamba model applied to VAD, designed to enable fast inference through efficient linear computation. Specifically, we developed VQ-MaU with a symmetric encoder-decoder structure, incorporating the NVSS blocks to accelerate convergence for normal features. Additionally, we introduced VQ layer to enhance feature retrieval and discrimination. To further improve anomaly detection, we integrated frame prediction and optical flow reconstruction, through a clip-level fusion evaluation strategy to ensure strong discrimination between appearance and motion. Extensive experiments on three datasets show that VADMamba demonstrates strong competitiveness while maintaining fast inference speeds compared to existing methods.

\bibliographystyle{IEEEbib}
\bibliography{refs}


\end{document}